%% file: main.tex
\documentclass{article}
\usepackage{amsmath,epsfig}
\usepackage{amssymb} 
\usepackage{multirow}
\usepackage{booktabs}
\usepackage{dsfont}
\usepackage{setspace}

\let\OLDthebibliography\thebibliography
\renewcommand\thebibliography[1]{
  \OLDthebibliography{#1}
  \setlength{\parskip}{0pt}
  \setlength{\itemsep}{0pt plus 0.3ex}
}
\usepackage[preprint]{spconf}
\copyrightnotice{978-1-6654-3864-3/21/\$31.00~\copyright2021 IEEE \hfill}

\begin{document}\sloppy

\def\x{{\mathbf x}}
\def\L{{\cal L}}

\title{Few-Shot Semantic Segmentation via Prototype Augmentation \\with Image-Level Annotations}
%

\name{Shuo Lei\textsuperscript{1}, Xuchao Zhang\textsuperscript{1,}\textsuperscript{2}, Jianfeng He\textsuperscript{1}\sthanks{Corresponding author.}, Fanglan Chen\textsuperscript{1}, Chang-Tien Lu\textsuperscript{1}}
\address{\textsuperscript{1}Department of Computer Science, Virginia Tech, Falls Church, VA, USA\\
\textsuperscript{2}NEC Laboratories America, Princeton, NJ, USA\\
\tt\small\{slei,xuczhang,jianfenghe,fanglanc,ctlu\}@vt.edu}


\maketitle

\begin{abstract}
Despite the great progress made by deep neural networks in the semantic segmentation task, traditional neural-network-based methods typically suffer from a shortage of large amounts of pixel-level annotations. Recent progress in few-shot semantic segmentation tackles the issue by only a few pixel-level annotated examples. However, these few-shot approaches cannot easily be applied to multi-way or weak annotation settings.
In this paper, we advance the few-shot segmentation paradigm towards a scenario where image-level annotations are available to help the training process of a few pixel-level annotations. Our key idea is to learn a better prototype representation of the class by fusing the knowledge from the image-level labeled data. Specifically, we propose a new framework, called PAIA, to learn the class prototype representation in a metric space by integrating image-level annotations. Furthermore, by considering the uncertainty of pseudo-masks, a distilled soft masked average pooling strategy is designed to handle distractions in image-level annotations. Extensive empirical results on two datasets show superior performance of PAIA.
\end{abstract}

\input{Introduction}
\input{Problemsetting}
\input{Model}
\input{Experiment}

\section{Conclusion}\label{section:conclusion}
In this paper, a novel weak-annotation-augmented few-shot segmentation model is proposed to learn an augmented prototype based on both pixel-level and image-level annotations. To achieve this, we design a robust strategy with soft-masked average pooling to handle the noise in image-level annotations. It considers the prediction uncertainty of the image-level annotations and employs the task-specific threshold to filter the distraction. Our evaluation results demonstrated the superiority of the proposed method over existing few-shot segmentation models by a sizeable margin.

\bibliographystyle{IEEEbib}
\small
\bibliography{Reference}

\end{document}

%% file: Introduction.tex
\section{Introduction}


Semantic segmentation, one of the most challenging tasks in computer vision, aims to assign a categorical label to each pixel of an image according to its enclosing object or region.
In the past few years, a number of deep-neural-network-based approaches have been proposed for the semantic segmentation task. However, training these models typically requires large-scale pixel-level annotations, which are expensive to obtain.
Some semi-/weakly-supervised segmentation models were proposed to reduce the dependence on pixel-level annotated data but still suffer from the issue of model generalization, which makes them hard to be applied to unseen categories. 

\begin{figure}[tb]
    \centering
    \includegraphics[width=\linewidth]{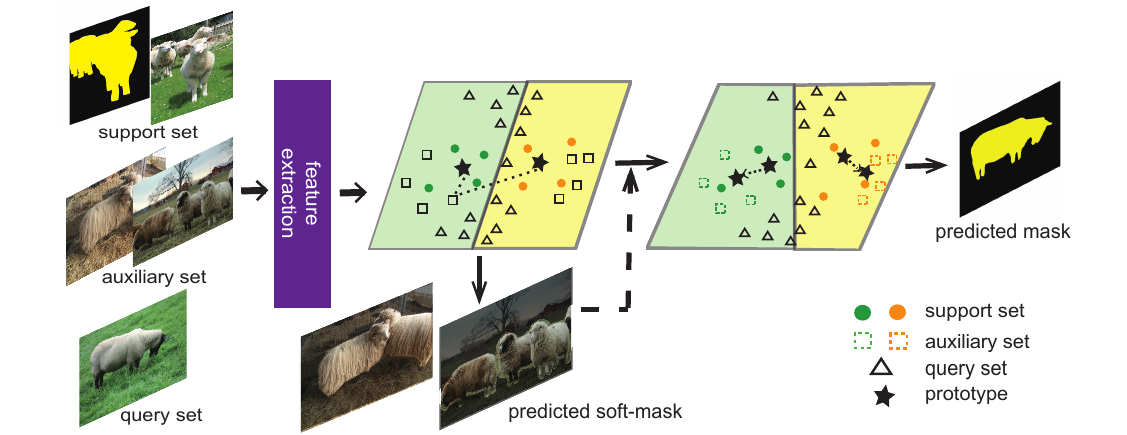}
    \caption{Illustration of a 1-way 1-shot with 2 image-level auxiliary images task. Our objective is to augment the class representation in the metric space by using image-level labeled data. Our model employs soft-mask average pooling strategy to handle the distraction in the image-level annotations and re-feed the related embedding features to refine the prototypes.}
    \label{fig:problem}
    \vspace{-0.3cm}
\end{figure}

Recently, there has been increasing interest in the study of few-shot semantic segmentation, which learns to perform segmentation on novel classes with only a few pixel-level annotated images. However, existing studies suffer from three issues: 1) Hard to handle the multi-way segmentation task. One group of methods~\cite{shaban2017one,zhang2019canet,zhang2019pyramid} perform segmentation by matching dense pair-wise features to query images. However, those methods mainly focus on one-way few-shot setting, and hence it is non-trivial to generalize them to the multi-way segmentation task. 2) Inaccurate class representations due to limited number of labeled data. The other group of studies~\cite{wang2019panet,liu2020prototype} are based on prototype learning, which learns a metric space to employ segmentation on new classes with similarity measurements. These prototype-based methods are highly dependent on the accuracy of prototype representations and provide a solution to the multi-way segmentation task. However, all those methods represent a semantic class based on a small support set, which are restrictive for capturing rich and accurate prototypes.
3) Poor robustness to the weak annotations. Recently, PPNet~\cite{liu2020part} adopts extra image-level annotations and decomposes the holistic prototypes to enrich the prototype representations. Although matching the part-aware prototypes captures fine-grained object features, it cannot handle the weak annotations (like scribble or bounding box annotations) in the support set since part-aware prototypes require accurate pixel-level annotations. In addition, it ignores the uncertainty of pseudo-labels when refining the prototypes. Concretely, it is non-trivial to select class-related parts in the image-level annotations due to lacking pixel-level annotations. Nevertheless, PPNet filters the irrelevant parts only by an unified pre-defined threshold, which may introduce some distractions to the prototypes.









To tackle the above issues, we first consider a new scenario to augment the prototype presentations, where the segmentation of novel classes are learned in the combination of pixel- and image-level annotations, which is shown in Fig.\ref{fig:problem}. Specifically, in a $N$-way $K$-shot segmentation task, we aim to perform segmentation on the query images with $K$ pixel-level and $U$ image-level annotated images from each of the $N$ classes. It is worth emphasizing that mixing strong (pixel-level) and weak (image-level) annotations is a widely used setting to improve the performance in the existing semantic segmentation works~\cite{luo2017deep,souly2017semi}. But these methods still require large weak annotations to guide the training process. Instead, the number $U$ of weak labeled data needed in our method is very small ($U<15$), which is more feasible for few-shot setting.
Second, we propose a distilled soft-masked average pooling strategy to handle the distraction in image-level annotations. It considers the uncertainty of the pseudo-mask and applies the task-specific threshold to filter the class relevant parts in image-level annotations, which is more effective than an unified pre-defined hard threshold. Finally, we propose an Iterative Fusion Module (IFM) to refine the prototypes by integrating the prototype of image-level annotations into original one. 

To sum up, our main contributions are as follows: 
\begin{itemize}
\setlength{\itemsep}{0pt}
\setlength{\parsep}{0pt}
\setlength{\parskip}{0pt}
    \item Propose a class-prototype augmentation method to enrich the prototype representation by utilizing a few image-level annotations, achieving superior performance in one-/multi-way and weak annotation settings.
    \item Design a robust strategy with soft-masked average pooling to handle the noise in image-level annotations, which considers the prediction uncertainty and employs the task-specific threshold to mask the distraction.
    \item Conduct extensive experiments on two datasets for performance evaluations. Our method outperforms the state-of-the-arts with less image-level annotations and can achieve 8.2\% and 6.8\% improvements in mIoU score for one-shot settings with scribble and bounding box annotations in PASCAL-$5^i$, respectively.
\end{itemize} 


%% file: Problemsetting.tex
\section{Problem Setting} \label{section:problem_setting}
Our purpose is that a model trained on a large labeled dataset $\mathcal{D}_{train}$ can make a segmentation prediction on a testing dataset $\mathcal{D}_{test}$ with a few annotated examples. The class set $\mathcal{C}_{train}$ in $\mathcal{D}_{train}$ has no overlap with $\mathcal{D}_{test}$, i.e., $\mathcal{C}_{test} \cap \mathcal{C}_{train} = \emptyset$. Following previous works~\cite{wang2019panet,zhang2019canet}, we adopt an episodic paradigm in the few-shot segmentation task. 
Each episode $e$ is composed by 1) a \textit{support} set $\mathcal{S}_e = \{(I_{i}^s,M_{i}^s)\}_{i=1}^{N\times K}$, containing $K$ $(image,mask)$ pairs for each of the $N$ categories in the foreground,
where $(I_{i}^s,{M}_{i}^s)$ represents the pair of support image $I_{i}^s$ and its corresponding binary mask $M_{i}^s$ for the foreground class, 
2) a \textit{query} set $\mathcal{Q}_e = \{(I_{j}^q,M_{j}^q)\}_{j=1}^{Q}$, which contains $Q$ different query sample pairs $(I_{j}^q,M_{j}^q)$ from the same $N$ categories, 
and 
3) an \textit{auxiliary} set $\mathcal{A}_e = \{I_{t}^a\}_{t=1}^{N\times U}$, containing $U$ image-level labeled images $I_{t}^a$ for each of the same $N$ categories, but no pixel-level annotation is available. The set of all target classes in the foreground for episode $e$ denotes as $\mathcal{C}_e$, and $|\mathcal{C}_e|=N$. 
For each episode $e$, the model is supposed to segment images from $\mathcal{Q}_e$ with the combination of $\mathcal{S}_e$ and $\mathcal{A}_e$. 

%% file: Model.tex
\section{Model}\label{section:model}

\subsection{Overall Architecture}
\begin{figure*}[htb]
    \centering
    \scalebox{0.8}{
    \includegraphics[trim=0cm 0cm 1cm 0.5cm]{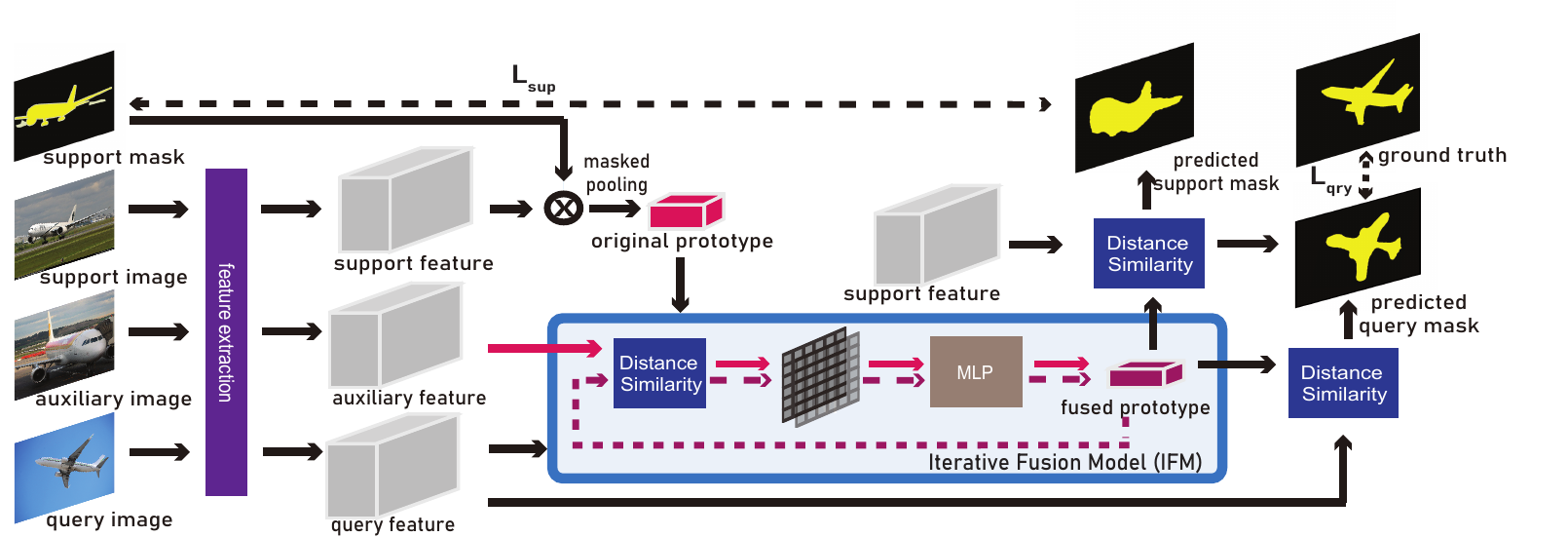}}
    \caption{Overview of the proposed network structure. The support, auxiliary, and query images are embedded into deep features via the feature encoder. Then the model employs masked average pooling over the support set to get the original prototypes. Iterative Fusion Module (IFM) augments the original prototypes by re-feeding the related auxiliary features with the proposed distilled soft-masked average pooling method. The fused prototypes are obtained via the IFM and are used to segment query images. $\mathcal{L}_{sup}$ and $\mathcal{L}_{qry}$ denote the standard cross entropy loss between the segmentation results and the ground truth of the support set and query set, respectively.}
    \label{fig:model}
\end{figure*}
We propose a new framework that can solve the few-shot segmentation problem with a combination of pixel- and image-level labeled data. The main idea of our model is to learn a better prototype representation of the class by fusing the knowledge from the image-level labeled data. Specifically, the original prototypes are first obtained on the support set and are used to segment image-level annotations. Class irrelevant parts in those images should be filtered and the most confidently related features are considered to augment the original prototypes.
To this end, we propose a novel prototype fusion strategy that contains the distilled soft-masked average pooling method and iterative fusion module, as shown in Fig. \ref{fig:model}. 
\subsection{Prototype Representation Learning}
We represent each category of segmentation task as a prototype in the metric space. The original class-specific prototypes are obtained by employing masked average pooling over the support set, which averages the features of the pixels only belonging to the support classes, and each pixel of the query image is labeled by its nearest prototype in the metric space. Thus, the prototype $p_c$ of the foreground class $c$ is defined as follows:
\begin{small}
\begin{align}
        p_c = \frac{1}{K}\left(\sum\mathrm{MP}(I_{c,i}^s,M_{c,i}^s)\right),
\end{align}
\end{small}
We have $I_{c,i}^s=\{x_j\}_{j=1}^{w\times h}$ and $M_{c,i}^s=\{y_j\}_{j=1}^{w\times h}$, where $w$ and $h$ denote the width and height of the image, respectively. $\mathrm{MP}(\cdot)$ denotes the masked average pooling function.
Moreover, the background prototype is computed by averaging all the features of the pixels that do not belong to any foreground class in $\mathcal{C}_e$.


\subsection{Soft-masked Average Pooling}
Our model enhances the prototypes by extracting more class representation knowledge from the additional image-level annotations. The most intuitive way to incorporate those image-level annotations is to obtain their pseudo masks and directly add them into the support set. However, this process may introduce some noise into the support set. Since the original prototypes can be biased due to support data scarcity, it may lead to inaccurate prediction results of image-level labeled data. To tackle this issue, we propose a soft-masked pooling method (SMP). Instead of assigning the same weight to each pixel belonging to the support class, we give them a partial assignment based on their probability of falling into the class. Pixels with lower predicted confidence would get lower weights preventing them from distracting the original prototypes.   
Specifically, for each foreground class $c$, we first compute the predicted probability map $Y_{c,i}^a=\{\hat{y}_j^c\}_{j=1}^{w\times h}$ and pseudo binary mask $\hat{M}_{c,i}^a=\{z_j\}_{j=1}^{w\times h}$ of $I_{c,i}^a$, where the indicator $z_j$ is set to 1 if $\hat{y}_j=c$. 
Then, we compute the representative vector by averaging the pixels within the object regions on the feature map. Thus, the soft-masked average pooling can be formed as: 
\begin{small}
\begin{align}
    \mathrm{SMP}(I_{c,i}^a,Y_{c,i}^a,\hat{M}_{c,i}^a) &= \frac{\sum_j f(x_j)\hat{y}_jz_j}{\sum_j\hat{y}_jz_j},
\end{align}
\end{small}where $f(\cdot)$ is defined as a feature encoder function. In this way, the original prototypes can be enhanced by incorporating part of image-level labeled samples. The fused prototypes can be computed as follows:

\begin{footnotesize}
\begin{equation}
    \label{eq:softproto}
    \tilde{p}_c = \frac{1}{K+U}(\sum_i\mathrm{MP}(I_{c,i}^s,M_{c,i}^s) 
    + \sum_j\mathrm{SMP}(I_{c,j}^a,Y_{c,j}^a,\hat{M}_{c,j}^a)).
\end{equation}
\vspace{-0.5cm}
\end{footnotesize}
\subsection{Distilled Soft-masked Average Pooling}
In our problem setting, each image contains at least one foreground class in an episode. Therefore, the categories of referred segmentation mask belong to at least two of the $N+1$ classes (including the background class). However, when computing the prototype of background class, we treat all pixels not belonging to foreground classes as the same category. That means we cannot guarantee two support images have similar background class representation even if their foreground classes are the same. Moreover, the image-level annotated image may contain an unseen object, which does not show up in the background or does not belong to any foreground classes in the support set.

Under this circumstance, those unlabeled pixels could still get pseudo labels with higher confidence even if they are far away from all prototypes in the metric space. So the uncertainty of these unseen objects in images may reduce the accuracy of fused prototypes. 
To alleviate the issue, we use a filter strategy for each prototype when applying soft-masked average pooling over those unlabeled images, which is called distilled soft-masked average pooling. Inspired by Ren et al.~\cite{ren2018meta}, we try to compute a threshold $\gamma_c$ based on the statistics of the distances between pixels and the prototypes. 
Specifically, we first compute the distance $d^c_j$ between the prototype $p_c$ and the pixel $j$, and obtain distance matrix $D_i^c$ of image $I_i^a$ for the prototype $p_c$, $i.e.$ $D_i^c=\{d_j^c\}_{j=1}^{w\times h}$. Then, normalized distance set $\tilde{D}^c$ is obtained by normalizing each distance from the distance set $D^c=\{D_i^c\}_{i=1}^{N\times U}$.
Finally, the filter threshold $\gamma_{c}$ for the prototype $p_c$ in each episode $e$ is defined as follows:

\begin{footnotesize}
\begin{equation}
    \gamma_{c} = \mathrm{MLP}([\mathop{\mathrm{min}}_{i}(\tilde{D}^{c}_{i}),\mathop{\mathrm{max}}_{i}(\tilde{D}^{c}_{i}),
    \mathop{\mathrm{var}}_{i}(\tilde{D}^{c}_{i}),
    \mathop{\mathrm{skew}}_{i}(\tilde{D}^{c}_{i}),\mathop{\mathrm{kurt}}_{i}(\tilde{D}^{c}_{i}]).
\end{equation}
\end{footnotesize}

For each foreground class $c$, the distraction indicator $\tau_i$ of pixel $i$ can be computed as $\mathds{1}(\mathrm{d}(f(x_i),p_c)<\gamma_{c})$, where $\mathds{1}(\cdot)$ is an indicator function, outputting value 1 if the argument is true or 0 otherwise. 
Then, the indicator $H_{c,i}^a=\{\tau_j\}_{j=1}^{w\times h}$ of $I_{c,i}^a$ is applied to filter the pixels that are not worth considering. 
In this way, the model is forced to only extract objective class-related pixels instead of considering the whole image which may contain novel object classes in the background. Therefore, Eq.\ref{eq:softproto} for the fused prototype computation can be updated as follows:

\begin{scriptsize}
\begin{equation}
    \label{eq:softprotomask}
    \tilde{p}_c = \frac{1}{K+U}(\sum_i\mathrm{MP}(I_{c,i}^s,M_{c,i}^s)
    + \sum_j\mathrm{SMP}(I_{c,j}^a,Y_{c,j}^a,H_{c,j}^a \odot \hat{M}_{c,j}^a)),
\end{equation}
\vspace{-0.4cm}
\end{scriptsize}where $\odot$ is the element-wise product.

\subsection{Iterative Fusion Module}
Intuitively, if the knowledge extracted from the image-level annotations in the auxiliary set can improve the performance of our model, we can also utilize the image-level annotated images from the query set.
As the original prototypes are inevitably biased due to data scarcity, the confidence of the initial probability maps of those images may not be high enough to be considered. Therefore, we iteratively repeat the refinement for several steps to optimize the fusion prototypes in the Iterative Fusion Module (IFM). This process is shown in Figure \ref{fig:model}. In particular, we first compute the probability maps via the original prototypes and re-feed the embedding features with distilled soft-masked average pooling to the IFM. Then we alternatively use fused prototypes in the last epoch to recompute the probability maps. In this way, the bias of original prototypes can be reduced by adding more class-related features from image-level annotations. The more accurate prototypes are, the higher confidence can be obtained and the more class-related features can be considered.







%% file: Experiment.tex
\vspace{-0.cm}
\section{Experiments}\label{section:experiment}
\subsection{Experimental Settings}
\textbf{Datasets.} We evaluated the performance of our model on two common few-shot segmentation datasets: PASCAL-$5^i$ and COCO-$20^i$. 
PASCAL-$5^i$ dataset is proposed by Shaban et.al~\cite{shaban2017one} and is created from PASCAL VOC 2012~\cite{everingham2010pascal} with SBD~\cite{hariharan2011semantic} augmentation. The 20 categories in PASCAL VOC are evenly divided into 4 splits, each containing 5 categories. We used the rest of the images that do not have segmentation labels but have category information in PASCAL VOC 2012 as the auxiliary set. Similarly, COCO-$20^i$ is built from MS COCO~\cite{lin2014microsoft} and 80 categories are split into 4 folds. 
As each image in MS COCO has its corresponding segmentation label, we used images in the validation folder as the auxiliary set. Models were trained on 3 splits and evaluated on the rest one in a cross-validation for both datasets. Following the same scheme for testing~\cite{wang2019panet}, we averaged the results from 5 runs with different random seeds, each run containing 1,000 episodes to get stable results. $N_{query} = 1$ is used for all experiments.



\textbf{Implementation details.} We adopted a VGG-16~\cite{simonyan2014very} and ResNet-50~\cite{he2016deep} network as the feature extractor following conventions. For the MLP used in the distilled soft-masked pooling, we used a single hidden layer with 20 hidden units with a Tanh activation.
For implementation, we used Pytorch~\cite{paszke2017automatic}. Following previous works~\cite{zhang2018sg,wang2019panet,nguyen2019feature}, we pretrained the CNN on ImageNet~\cite{russakovsky2015imagenet}. All images were resized to $417\times417$ and augmented by random horizontal flipping. The network was trained end-to-end by SGD with a learning rate of 1e-3, momentum of 0.9 and weight decay of 5e-4. We trained the model in 20,000 iterations and the batch size is 1. The learning rate was reduced by 0.1 after 10,000 iterations. We adopted mean-IoU as the metric method to evaluate the model performance~\cite{wang2019panet,zhang2019canet}. PANet* denotes the baseline that taking ResNet-50~\cite{he2016deep} as feature extractor in PANet~\cite{wang2019panet} .

\begin{table*}[htb]
\small
\begin{center}
\caption{Mean-IoU of 1-way 1-shot and 5-shot segmentation on PASCAL-$5^i$. $S^u$ denotes the image-level annotations. CANet reports multi-scale test performance. The single-scale test performance is reported by~\cite{yang2020prototype}.}
\begin{tabular}{l|c|c|ccccc|ccccc}
\toprule
\multirow{2}{*}{Methods} & \multirow{2}{*}{$S^u$} & \multirow{2}{*}{Backbone} & \multicolumn{5}{c|}{\bf 1-shot}                   & \multicolumn{5}{c}{\bf 5-shot}                   \\
            &     &    & split-1 & split-2 & split-3 & split-4 & mean & split-1 & split-2 & split-3 & split-4 & mean \\
\hline
OSLSM~\cite{shaban2017one} & $\times$ & VGG16        & 33.60    & 55.30    & 44.90    & 33.50    & 40.80 & 35.90    & 58.10    & 42.70    & 39.10    & 43.90 \\
SG-One~\cite{zhang2018sg} & $\times$ & VGG16  & 40.20    & 58.40    & 48.40    & 38.40    & 46.30 & 41.90    & 58.60    & 48.60    & 39.40    & 47.10 \\
AMP~\cite{siam2019amp} & $\times$   & VGG16     & 36.80    & 51.60    & 46.90    & 36.00    & 42.80 & 44.60    & 58.00    & 53.30    & 42.10    & 49.50 \\
FWB~\cite{nguyen2019feature} & $\times$  & VGG16   & 47.04    & 59.64    & 52.61    & 48.27  & 51.90  & 50.87 & 62.86    & 56.48    & 50.09    & 55.08  \\
PANet~\cite{wang2019panet} & $\times$ & VGG16    & 42.30    & 58.00    & 51.10    & 41.20    & 48.10 & 51.80    &64.60    & 59.80    & 46.50    & 55.70 \\
PANet*~\cite{wang2019panet}  & $\times$ & RN50    & 44.03    & 57.52    & 50.84    & 44.03    & 49.10 & 55.31    &67.22    & 61.28    & 53.21    & 59.26 \\
CANet~\cite{zhang2019canet}  & $\times$ & RN50    & 49.56    & 64.97    & 49.83    &\bf 51.49    & 53.96 & -    & -    & -    & -    & 55.80 \\
PMMs~\cite{yang2020prototype} & $\times$ & RN50    & \bf 51.98    & \bf 67.54    & 51.54    & 49.81    & \bf 55.22 & 55.03    & 68.22    & 52.89    & 51.11    & 56.81 \\
\hline
PPNet~\cite{liu2020part} & $\checkmark$ & RN50    & 48.58    & 60.58    & 55.71    & 46.47    & 52.84 & 58.85    & 68.28    & 66.77    & 57.98    & 62.97 \\
\hline
\bf PAIA (ours) & $\checkmark$  & VGG16  & 49.50    & 61.64    & \bf 56.03    & 45.61    & 53.20 & 54.14    & 63.82    & 62.30    & 50.53    & 57.70 \\
\bf PAIA (ours) & $\checkmark$  & RN50  & 50.31    & 62.10    & 55.97    & 47.72  & 54.03  &\bf 59.70    & \bf 69.82    & \bf 66.92    &\bf 59.73    &\bf 64.04 \\
\bottomrule
\end{tabular}
\label{exp:1waymiou}
\end{center}
\vspace{-0.3cm}
\end{table*}

\begin{table*}[htb]
\begin{center}
\small
\caption{Mean-IoU of 2-way 1-shot and 5-shot segmentation on PASCAL-$5^i$. $S^u$ denotes the image-level annotations.}
\begin{tabular}{l|c|c|ccccc|ccccc}
\toprule
\multirow{2}{*}{Methods} & \multirow{2}{*}{$S^u$} &  \multirow{2}{*}{Backbone} & \multicolumn{5}{c|}{\bf 1-shot}                   & \multicolumn{5}{c}{\bf 5-shot}                   \\
                &    &     & split-1 & split-2 & split-3 & split-4 & mean & split-1 & split-2 & split-3 & split-4 & mean \\
\hline
SG-One~\cite{zhang2018sg} & $\times$ & VGG16    & -       & -       & -       & -       & -    & -       & -       & -       & -       & 29.40 \\
PANet~\cite{wang2019panet} & $\times$ & VGG16       & -       & -       & -       & -       & 45.10 & -       & -       & -       & -       & 53.10 \\
\hline
PPNet~\cite{liu2020part} & $\checkmark$ & RN50    & 47.36   & 58.34    & 52.71 & \bf48.18   & 51.65 & \bf 55.54 & 67.26  & 64.36   & 58.02 & 61.30 \\
\hline
\bf PAIA (ours) & $\checkmark$ & VGG16      & 47.71    & \bf 60.04    & 53.30    & 46.02      &  51.76 & 49.10    & 60.72    & 58.64    & 48.22    & 54.17\\
\bf PAIA (ours) & $\checkmark$ & RN50     & \bf 48.94    & 59.95    & \bf 54.11    & 47.79      & \bf 52.70 & 55.21    & \bf68.50    & \bf64.97    & \bf59.76    &\bf 62.11\\
\bottomrule
\end{tabular}
\end{center}
\label{exp:2way}
\vspace{-0.3cm}
\end{table*}

\subsection{Comparison with the Competing Methods}
We first compared our PAIA model with the state-of-the-art methods on PASCAL-$5^i$ dataset in 1-way segmentation task. Table \ref{exp:1waymiou} shows the results in mean-IoU metric. Specifically, compared with PANet~\cite{wang2019panet}, our model achieves an improvement of $\bf 4.93\%$ in the 1-way 1-shot task and $\bf 4.78\%$ in the 5-shot task, which means the combination of both pixel-level and image-level annotations can improve the performance in the few-shot segmentation task. Moreover, our model employs $\bf2$ image-level annotations and surpasses PPNet~\cite{liu2020part} by $\bf1.19\%$ for 1-shot and $\bf1.07\%$ for 5-shot while the PPNet method even uses $\bf 3$ times more image-level annotations. This indicates that our soft-masked average pooling strategy can enhance the utilization of the image-level annotations. 
Table \ref{exp:coco} shows the evaluation results on COCO-$20^i$. Compared to PASCAL VOC dataset, MS COCO has more object categories, which makes it more difficult than PASCAL VOC. Our model can outperform the previous methods due to the capability of extracting class-related knowledge from the image-level annotations even though more unseen objects are included. 
\begin{figure}[htb]
    \centering
    \scalebox{0.85}{
    \includegraphics[width=\linewidth]{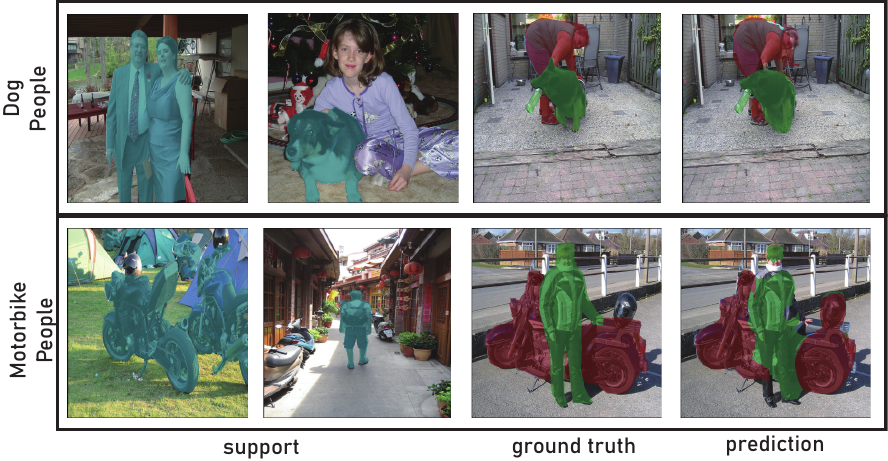}}
    \caption{Qualitative examples of 2-way 1-shot segmentation on the PASCAL-$5^i$.}
    \label{fig:2way}
\end{figure}
\vspace{-0.7cm}

\begin{table}[tb]
\begin{center}
\small
\setlength{\tabcolsep}{3pt}
\caption{Mean-IoU of 1-way segmentation on COCO-$20^i$. $S^u$ denotes the image-level annotations.}
\begin{tabular}{l|c|c|cc}
\toprule
\multirow{2}{*}{Method} & \multirow{2}{*}{$S^u$} & \multirow{2}{*}{Backbone} & \multicolumn{2}{c}{mean-IoU} \\ \cline{4-5} 
        &   &   & \bf 1-shot        & \bf 5-shot    \\ \hline
PANet*~\cite{wang2019panet}  &   $\times$  & RN50        & 22.95         & 33.80      \\ \hline
PPNet~\cite{liu2020part}   &  $\checkmark$ & RN50  & 27.16      & 36.73      \\ \hline
\bf PAIA (ours)    &$\checkmark$  & RN50  & \textbf{28.12}        & \textbf{37.63}  \\ \bottomrule
\end{tabular}
\label{exp:coco}
\vspace{-0.7cm}
\end{center}
\end{table}

\begin{figure}[htb]
    \centering
    \scalebox{0.85}{
    \includegraphics[width=\linewidth]{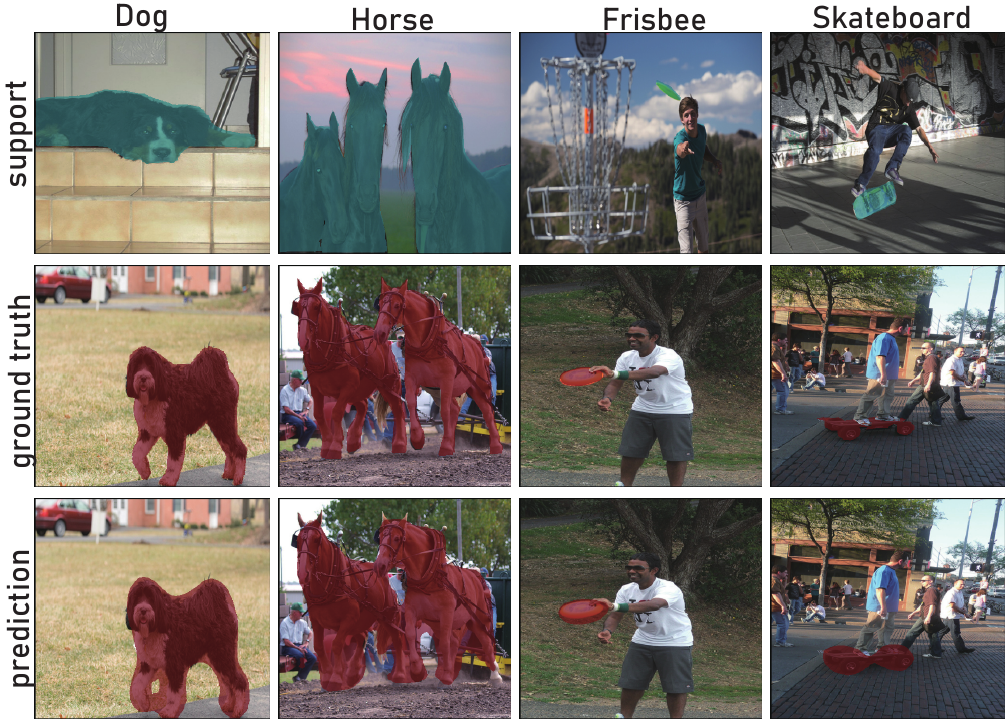}}
    \caption{Qualitative examples of 1-way 1-shot segmentation. The left two columns are on PASCAL-$5^i$ and the right two columns are on COCO-$20^i$.}
    \label{fig:exp1shot}
    \vspace{-0.3cm}
\end{figure}

\subsection{Results on Weak Annotations}
To validate the robustness of our model, we evaluated PAIA with two types of weak annotations: scribble and bounding box. The pixel-level annotations of the support set were replaced by scribbles or bounding boxes. For fair comparison, we adopted VGG-16 as feature extractor and used the same annotation settings in Wang et al.~\cite{wang2019panet}: scribbles are generated from the dense segmentation masks automatically and bounding box is randomly chosen from instance mask.
\begin{table}[htb]
\small
\setlength{\tabcolsep}{3.5pt}
\begin{center}
\caption{Mean-IoU of using different types of annotations on PASCAL-$5^i$. $S^u$ denotes the image-level annotations.}
\begin{tabular}{l|c|cc|cc|cc}
\toprule
\multirow{2}{*}{Method} & \multirow{2}{*}{$S^u$} & \multicolumn{2}{c|}{Scribble}    & \multicolumn{2}{c|}{Bounding Box} & \multicolumn{2}{c}{Densed}                \\ \cline{3-8} 
                    &    & 1-shot        & 5-shot        & 1-shot         & 5-shot        & 1-shot        &       5-shot        \\ \hline
PANet~\cite{wang2019panet}  &$\times$         & 44.8          & 54.6          & 45.1           & 52.8          & 48.1          &          55.7          \\ \hline
\bf PAIA   & $\checkmark$   & \textbf{53.0} &  \textbf{57.3} & \textbf{51.9}  & \textbf{56.2} & \textbf{53.2} &  \textbf{57.7}   \\ \bottomrule
\end{tabular}
\label{exp:weak}
\vspace{-0.7cm}
\end{center}
\end{table}

As shown in Table \ref{exp:weak}, for scribble annotations, our model achieves significant improvements of $\bf 8.2\%$ and $\bf 2.7\%$ in 1-shot and 5-shot tasks, respectively. This performance is comparable to the result with an expensive pixel-level annotated support set, which means our model works very well with sparse annotations.
In addition, with bounding box annotations, our model significantly outperforms the state-of-the-art methods by $\bf 6.8\%$ for the 1-shot task and $\bf 3.4\%$ for the 5-shot task.
This demonstrates that our model has a greater ability to withstand the noise introduced by the background area within the bounding box. Furthermore, the improvements in weak annotations validate the robustness of our model. 
\subsection{Ablation Study}
We implemented extensive ablation experiments on the PASCAL-$5^i$ dataset to evaluate the effectiveness of different components in our network by using the mean-IoU metric in the 1-way 1-shot task. In Table \ref{exp:ablation}, we compare our model with two baseline models. The first one does not adopt the distilled strategy when applying soft-masked pooling (DSMP), which is denoted as PAIA-Smp. The second one does not employ an additional iterative fusion module for the fused prototypes, i.e., the initial prediction from PAIA (PAIA-Init). As shown in Table \ref{exp:ablation}, the distilled soft-masked pooling method achieves a 2.2\% improvement over the soft-masked pooling method. In addition, the iterative fusion module yields an improvement of 1.1\% over the initial prediction. The combination of both modules achieves the best performance.
\vspace{-0.3cm}
\begin{table}[htb]
\begin{center}
\small
\caption{Ablation study on the choice of proposed module on 1-way 1-shot segmentation task on PASCAL-5$^i$.}
\begin{tabular}{c|c|ccccc}
    \toprule
        & Backbone & DSMP &  & IFM &  & mean-IoU      \\ \hline
    PAIA-Smp & VGG16 &  $\checkmark$    &  &     &  & 51.0          \\ \hline
    PAIA-Init & VGG16 &     &  &  $\checkmark$   &  & 52.1          \\ \hline
    PAIA    & VGG16 &  $\checkmark$    &  &  $\checkmark$   &  & \textbf{53.2} \\ \bottomrule
\end{tabular}
\vspace{-1cm}
\label{exp:ablation}
\end{center}
\end{table}


